\documentclass[10pt, a4paper]{article}
\usepackage{graphbox}
\usepackage{amsmath}
\usepackage{amssymb}
\usepackage{enumitem}

\usepackage{lrec-coling2024} 

\title{Towards a Generative Approach for Emotion \\Detection and Reasoning}

\name{Ankita Bhaumik, Tomek Strzalkowski} 

\address{Rensselaer Polytechnic Institute \\
         Troy, New York \\
         \{bhauma, tomek\}@rpi.edu\\}

\abstract{
Large language models (LLMs) have demonstrated impressive performance in mathematical and commonsense reasoning tasks using chain-of-thought (CoT) prompting techniques. But can they perform emotional reasoning by concatenating `Let's think step-by-step' to the input prompt? In this paper we investigate this question along with introducing a novel approach to zero-shot emotion detection and emotional reasoning using LLMs.
Existing state of the art zero-shot approaches rely on textual entailment models to choose the most appropriate emotion label for an input text. We argue that this strongly restricts the model to a fixed set of labels which may not be suitable or sufficient for many applications where emotion analysis is required.
Instead, we propose framing the problem of emotion analysis as a generative question-answering (QA) task. Our approach uses a two step methodology of generating relevant context or background knowledge to answer the emotion detection question step-by-step.
Our paper is the first work on using a generative approach to jointly address the tasks of emotion detection and emotional reasoning for texts.
We evaluate our approach on two popular emotion detection datasets and also release the fine-grained emotion labels and explanations for further training and fine-tuning of emotional reasoning systems.
 \\ \newline  }

\begin{document}

\maketitleabstract

\section{Introduction}

Emotion detection from text has been a long standing research problem in NLP, and has constantly advanced with the development of newer tools and technologies. Traditional emotion detection systems used lexicons and machine learning algorithms to predict the emotion label for an input text. With the availability of large annotated datasets, training deep learning systems and pre-trained models became popular to address this task. All such existing methods approach this problem as a text classification task, aiming to select one or more emotion labels from a predefined set of labels. However most of these emotion label sets are very basic and restrictive. For example, popular Ekman emotions like \textit{happiness} and \textit{sadness} are generalized and do not convey much context specific information in any situation. \citep{coppini2023experiments} perform several real-life experiments to demonstrate the limitations of widely used categorical emotion models in every day situations.
Current state of the art zero-shot emotion detection systems prompt textual entailment models using the the fixed set of emotion labels for a dataset \citep{yin2019benchmarking}. Thus, the model needs to be prompted for every label to determine the one with the highest entailment score. Further, suitable prompts need to be designed for every dataset to achieve optimal performance \citep{basile-etal-2021-probabilistic, plaza-del-arco-etal-2022-natural}. This further demonstrates the need of a more generalized approach that works across different corpora and domains.

\begin{figure}[!t]
\centering \includegraphics[width=\linewidth]{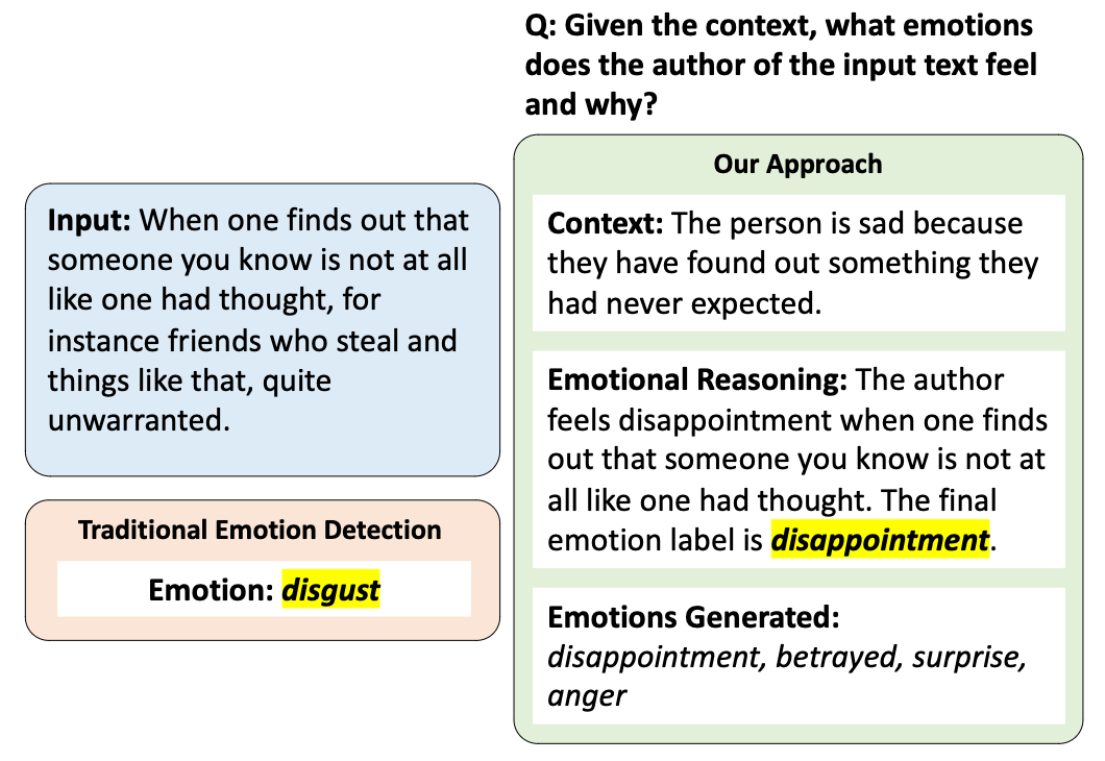}
    \caption{Example input text taken from ISEAR dataset. Our approach generates an open ended set of emotions along with an emotional reasoning for a final answer.}
    \label{fig:intro}
    \vspace{-0.1in}
\end{figure}

As humans we naturally follow a reasoning process to analyze the emotions in a situation. This process is influenced by diverse elements like background information, external stimuli, commonsense knowledge, and personal experiences.
Similarly, while analyzing the emotions in a piece of text we instantaneously perform this reasoning before attaching a final emotion label to it. Our work is inspired by this idea of generating a suitable emotion label following by a step-by-step reasoning process (Fig. \ref{fig:intro}). We introduce the novel idea of framing the emotion detection problem as a generative QA task. We leverage the power of LLMs that have been used popularly to solve arithmetic and commonsense reasoning tasks \citep{brown2020language, wei2022chain, narang-self}. We further investigate whether these models can perform this step-by-step emotional reasoning using CoT prompting techniques. Our preliminary experiments show that while the large LLMs having over 175B parameters manage to provide a reasoning, the task is notably more challenging for their smaller counterparts. 

Our approach first prompts the LLM to generate necessary background information or corpus specific context to answer the emotion detection question. It then utilizes the generated context to prompt the model to generate a step-by-step reasoning and a final emotion label. We also use the same prompt to perform traditional emotion detection over the fixed set of emotion labels available for the corpus. The probabilities of the emotion labels are sampled to predict the most probable one among them. Finally, we run our selection algorithm over the set of generated explanations and labels to select the most consistent ones using soft-majority voting. Due to the lack of availability of emotion reasoning datasets, we evaluate our approach over popular emotion detection datasets. We carry out manual evaluations over the generated explanations and release the annotated datasets for further use. 

Our generative approach has several advantages over traditional emotion detection systems.
Firstly, we move away from using a restrictive set of emotion labels to generating a more flexible open ended set of emotions that would be more suitable for the corpus. For example, for the input situation, \textit{"When I think about the short time that we live and relate it to  the periods of my life when I think that I did not use this short time"}, the gold label is \textit{sadness}. However, when the model attempts to reason step-by-step it generates the label \textit{regret}, which is more apt.
Further, the integration of context generation into the emotion detection process enables more corpus specific emotion analysis. It aids the model to perform in context learning and produce a relevant chain of thought for the emotional reasoning process. 
Finally, the generated explanation for the emotion labels makes the system explainable and adds value to the emotion analysis task.

Overall the contributions of our paper are as follows:
\begin{enumerate}[itemsep=0.05ex, topsep=1pt]
    \item A novel generative approach to emotion detection from text
    \item Demonstration of the significance of context incorporation into emotion detection prompts
    \item A methodology to generate CoT explanations along with emotion labels for an input text
    \item A selection algorithm to retrieve the top-k explanations from the model outputs
    \item An updated version of existing emotion datasets (ISEAR and \#Emotional Tweets) with the integration of open ended emotion labels and explanations
\end{enumerate}


\section{Emotional Reasoning}
We introduce the task of generating an emotional reasoning along with the predicted emotion label for an input text. This reasoning can be described as a step-by-step thought process that would lead us to finally predict one emotion label for the input. 
For example, when someone says \textit{"The weather is so gloomy today"}, we immediately reason that gloomy weather makes people \textit{sad} or \textit{unhappy}. 
However, for inputs which do not have an explicit cause mentioned like \textit{"The new president will stop the construction of the wind turbines."}, we would require further context and the author's perspective to understand the emotions that they are trying to express. It could either be \textit{happiness} or \textit{disappointment} depending on the context of the situation \citep{lee-etal-2023-empathy}. This further highlights how emotional reasoning with context incorporation is significantly different from previous work on emotion cause extraction \citep{chen2010emotion, chen2018joint, xia-ding-2019-emotion}.

Instead of creating new datasets for the task of emotional reasoning, we utilize the ISEAR and \#Emotional Tweets datasets that are already annotated for emotion labels. As these datasets have shown promising results for fine-tuning on emotion classification, we believe that introducing explanations alongside these labels would enhance the potential for building systems that are more interpretable and explainable.
We present updated versions of these datasets that include fine-grained emotion labels that express the emotions of the author more closely than the gold labels, and top-3 explanations for these labels.

\begin{figure*}[!t]
\centering
    \includegraphics[width=\linewidth]{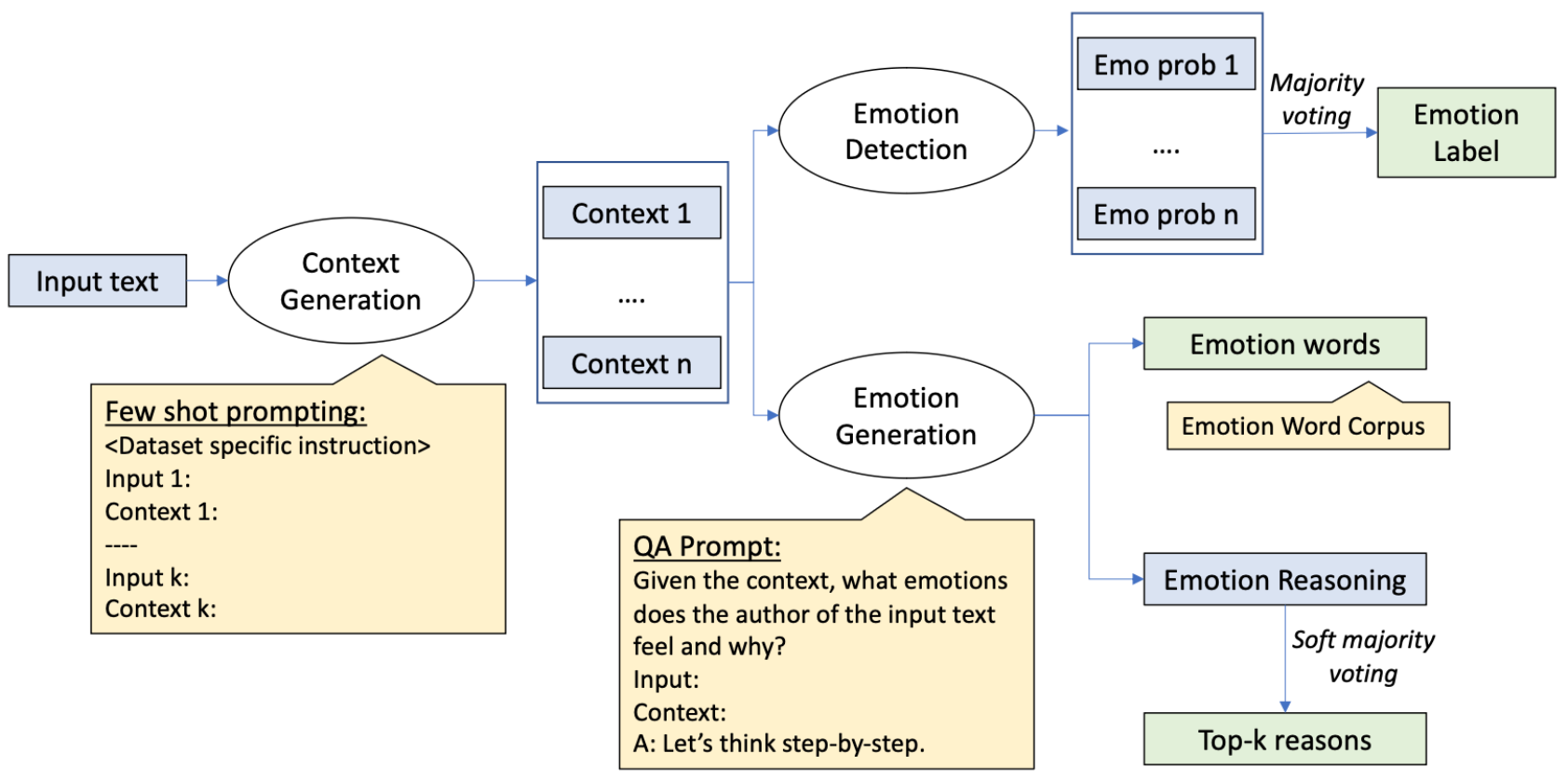}
    \caption{Overall architecture of our approach to generate: (1) an emotion label over a fixed set of labels (2) an open ended set of emotion words (3) top-k explanations}
    \label{fig:approach}
    \vspace{-0.2in}
\end{figure*}


\section{Methodology}

We present a generative approach to perform zero-shot emotion detection and reasoning over textual input. The overall architecture is illustrated in Fig. \ref{fig:approach}. First, we perform context generation to retrieve necessary background context for the specific domain and dataset. Next, we use this context along with a QA prompt to either detect the most probable emotion from a fixed set of labels or generate a list of suitable emotion words and a step-by-step reasoning for each label. For this paper, we focus only on a single step of reasoning for each input.

\subsection{Context Generation}
The context generation step uses few-shot prompting to extract the relevant context from a LLM. We provide a few handwritten examples that provides the LLM background about the application domain and the dataset construction process. For example, the ISEAR dataset reports situations in which people had experienced an emotion $e$. On the other hand the \#Emotional Tweets dataset \citep{mohammad-2012-emotional} collects tweets using emotion labels as hashtags. Our handwritten few shot prompt includes such background information to guide the model 
to generate the type of context that would aid in the emotion detection task.

The prompt ($P$) for context generation consists of an instruction, followed by $k$ pairs of input and relevant context for that dataset. The prompt for context generation can be constructed as follows:
\vspace{-0.1in}
\noindent\begin{align*}
    Prompt(P) &= {<}\texttt{instruction}{>} \\
               & \texttt{Input}: {<}i_1{>} \\ 
               & \texttt{Context}:{<}c_1{>} \\
               & ... \\
               & \texttt{Input}: {<}i_k{>} \\
               & \texttt{Context}:{<}c_k{>} \\
               & \texttt{Input}: {<}i_{k+1}{>} \\
               & \texttt{Context}:
\end{align*}

We generate $C = \{c_1, c_2 ... , c_{n}\}$ , a set of $n$ contexts for each input text. An example context generation prompt is shown in Table \ref{tab:isear_context}.

\begin{table*}[h]
\small
    \centering
    \begin{tabular}{p{0.9\linewidth}}
    \hline
    Generate the context for the situation described in the input.\\
Here are some examples: \\ \\

Input: I did not do the homework that the teacher had asked us to do. I was scolded immediately.\\
Context: This situation suggests that the person is a student who did not complete their homework as instructed by their teacher. \\ \\

Input: My parents were out and I was the eldest at home.  At midnight a male stranger phoned us and spoke to me in a rough language. I hung up and heard someone walking outside our door. \\
Context: This situation describes a home invasion or a potential break-in which could have been very frightening for the person at home, and may have left them feeling vulnerable and afraid.\\

...

Input: \{input text\}\\
Context:\\
        \hline
    \end{tabular}
    \vspace{-0.1in}
    \caption{Prompt for context generation for ISEAR dataset.}
    \label{tab:isear_context}
\end{table*}

\subsection{Emotion Generation}
Due to the huge success of LLMs in various QA tasks with CoT, we frame the emotion detection task in a similar format. We use the generated context as a part of this prompt to provide the LLM additional background knowledge about the input text and the dataset. The prompt $Q$ is constructed for emotion detection over a fixed set of labels $E$ and for open-ended generations. \\

\noindent \textit{Prompt(Q)} = \\
\noindent \texttt{Q: Given the context, what emotions does  the author of the input text feel and why?\\
Give me the reason followed by the final emotion label.\\
Context: \{context\}\\
Input: \{input\}\\
A: Let's think step-by-step.}\\

To predict the emotion of an input text $t$ over a finite set of labels $E$, we prompt the LLM $m$ to obtain the probability of every $e \in E$. We complete the prompt $Q$ 
using each context $c_i \in C$ to generate the most probable emotion label $\hat{e_i}$:

\begin{equation}
    \hat{e_i} = \arg \max_{e \in E}{p_m(e | t, c_i)}
\end{equation}

We obtain a set of $n$ emotion predictions for every context as $\hat{E} = \{(c_1,\hat{e}_1), (c_2,\hat{e}_2) ... (c_n,\hat{e}_n)\} $. Following earlier work on self-consistency \citep{narang-self}, we apply a marginalization over $\hat{E}$ to select the most consistent emotion label $\mathbf{e}$ using majority voting:

\begin{equation}
    \mathbf{e} = \arg \max_{i}{\sum_{i=1}^{n}{1(\hat{e_i} = \mathbf{e})}}
\end{equation}

To generate other suitable emotion words and corresponding explanations, we simply use nucleus sampling to complete the emotion QA prompt. Conditioning the generated output with the addition of context helps us to generate a suitable reasoning for that particular domain/dataset.
We sample the model outputs for every context $c_i \in C$ to produce a set of $q$ explanations.


\subsection{Answer Selection}
In the final step we design an answer selection algorithm to retrieve all emotion words and the top-k emotional reasoning answers from the $q$ generated outputs. Due to the open ended nature of LLM generations, this step is essential in eliminating any arbitrary or incomplete outputs that may have been generated.
We only select outputs that have been generated with a very high confidence. To implement this idea we use a soft majority voting technique that finds the most common outputs using semantic similarity rather than relying solely on exact matches. We use BERTScore, a popular evaluation metric for text generation to measure semantic similarity between the output texts \citep{zhang2019bertscore}. This uses contextual embeddings to represent the tokens
and compute agreement using cosine similarity.
Further, we use an external corpus of emotion/feeling words\footnote{\url{https://www.berkeleywellbeing.com/list-of-emotions.html}} to extract only emotion words from the top-k generations.

\section{Experiments}

\subsection{Datasets}
To evaluate our approach we choose datasets from two different domains that have been popularly used for existing works on zero-shot emotion detection.

\noindent \textbf{ISEAR} (International Survey on Emotion Antecedents and Reactions) contains 7,665 texts where participants reported situations when they felt the emotions of anger, disgust, fear, joy, sadness, shame and guilt \citep{scherer1994evidence}.\\

\noindent \textbf{\#Emotional Tweets} consists of 21,051 tweets that were collected using hashtags corresponding to the emotion labels anger, disgust, fear, happy, sadness, and surprise \citep{mohammad-2012-emotional}.

\subsection{Implementation}
We use Flan-T5 base (250M) and Flan-T5 xxl (11B) as our text generation models. 
The few-shot prompt $P$ is composed of $k=5$ hand-written samples. For context generation, we use nucleus sampling with $p$ set to 0.9 to generate $n=10$ contexts.
To generate an open-ended emotion label/reasoning we again use nucleus sampling with $p$ as to 0.9 to generate 10 explanations for every context.
The maximum number of tokens generated at every step can be 60.
Our experiments have been performed on 2 NVIDIA A100 GPUs.

\subsection{Baselines}

\textbf{Standard Zero-shot Prompting}: We use a straightforward prompt that is used for most text classification tasks, and greedy decoding generate the predicted emotion label: \\
\texttt{
This is an emotion classification task. \\
Text: \{input\} \\
Emotion: \\
}


\noindent \textbf{Prompting using CoT}: We append \textit{Let's think step-by-step} to the standard QA prompt and use greedy decoding to predict the emotion label: \\
\texttt{
Q: What emotion is expressed by the author in the input text? Let's think step-by-step  \\
Text: \{input\} \\
Emotion: \\
}

\noindent \textbf{State-of-the-art Zero-shot Emotion Detection}: The SOTA zero-shot classifiers use Natural language inference models to predict the textual entailment score for each emotion label. We report scores from the latest work by \citealp{plaza-del-arco-etal-2022-natural} where the authors experiment across various prompts and report the best performance with their ensemble models.

\section{Results}

\subsection{Zero-shot Emotion Detection}
We report the evaluation metrics for zero-shot emotion detection over the fixed set of emotion labels in both the datasets. Though the novel contribution of our method is to generate an open ended set of emotion labels and explanations, we report these scores to highlight that our approach would be effective on all three emotion analysis tasks.
We observe that using simple prompting techniques or CoT results in very poor performance in the task of emotion classification. However, using a QA prompt like ours (EmoGen) boosts the performance and thus can be used in place of existing textual entailment models for emotion detection.

\begin{table}[!h]
\small
    \centering
    \begin{tabular}{l|cc|cc}
        \hline
        & \multicolumn{2}{|c|}{ISEAR} & \multicolumn{2}{c}{\#Emo} \\
        \hline
        & Acc   & F1 & Acc   & F1\\
        \hline
        \textbf{\textit{Regular Prompting}} & & & &\\
        \hline
        FlanT5-base              & 0.17 & 0.25 & 0.06 & 0.13  \\
        FlanT5-xxl               & 0.39 & 0.45 & 0.09 &  0.17\\
        \hline
        \textbf{\textit{CoT Prompting}} & & \\ 
        \hline
        FlanT5-base              & 0.08 & 0.14 & 0.03 & 0.06 \\
        FlanT5-xxl               & 0.12  & 0.20 & 0.08 & 0.15\\
        \hline
        \textbf{\textit{NLI SOTA}} & - & 0.59 & - & 0.41 \\
        \hline
        \textbf{\textit{Emo Gen}} & & \\
        \hline
        FlanT5-base             & 0.52 & 0.50 & 0.40 & 0.32 \\
        FlanT5-xxl              & \textbf{0.62} & \textbf{0.60} & \textbf{0.48} & \textbf{0.42} \\
        \hline
    \end{tabular}
    \vspace{-0.1in}
    \caption{Accuracy and macro-F1 scores on the task of emotion classification. The highest scores for each dataset are in \textbf{bold}.}
    \label{tab:results}
    \vspace{-0.1in}
\end{table}

\subsection{Emotional Reasoning Dataset}
\begin{figure*}[t]
  \centering
  \includegraphics[width=.49\textwidth, align=t]{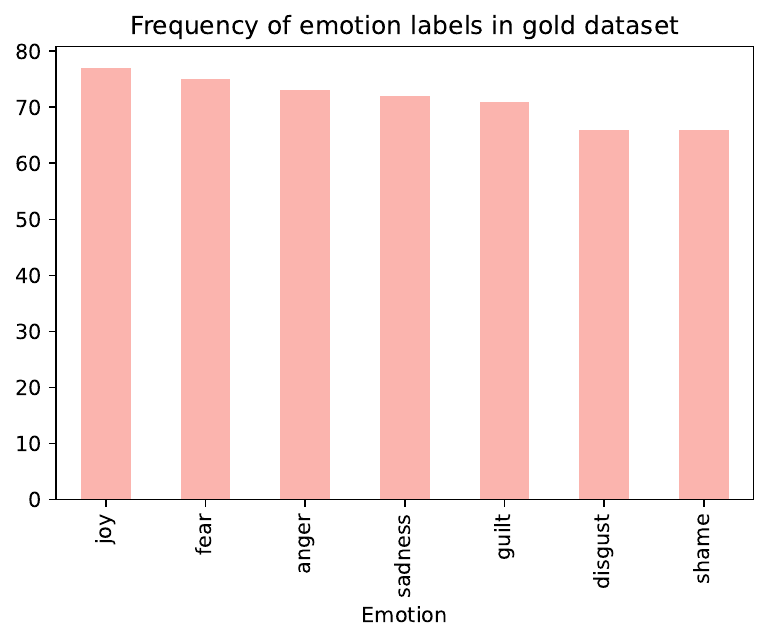}  
  \label{fig:isear_gold_dist}
  \includegraphics[width=.49\textwidth, align=t]{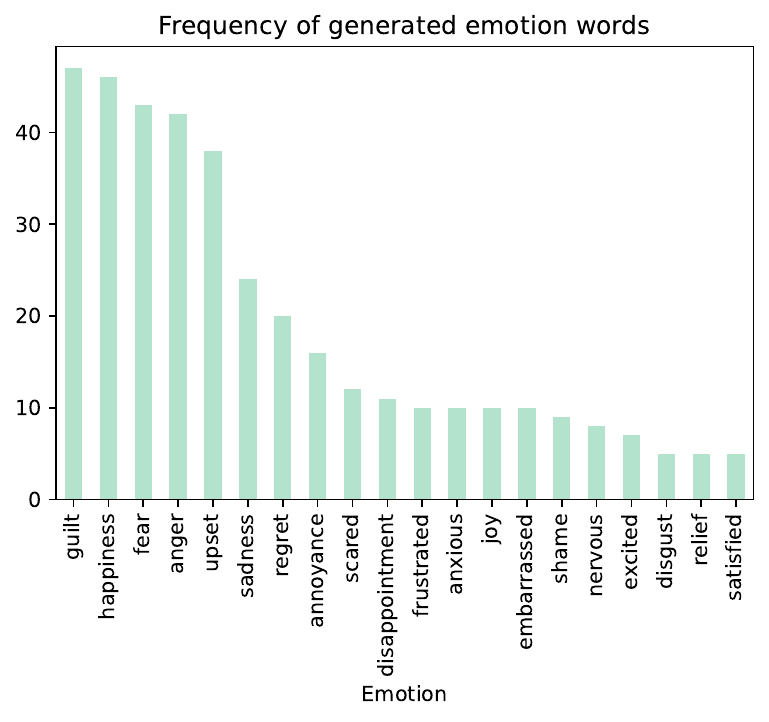}  
  \label{fig:isear_gen_dist}
\caption{Comparison of distributions of emotion words in gold ISEAR dataset vs. generated emotion labels using our approach}
\label{fig:distributions}
\vspace{-0.1in}
\end{figure*}

We generate updated versions of both the datasets to include the top-3 emotion labels beyond the predefined list of labels in the gold datasets. We also include an emotional reasoning for each label that replicates the chain-of-thought that the model used to generate the final label. We plot the frequencies of emotion labels in 500 samples of the ISEAR dataset to compare the distributions of gold emotion labels and generated labels. 
The distributions in Fig. \ref{fig:distributions} illustrate the breakdown of certain emotion labels into more detailed sub-labels.
For example, \textit{sadness} has been further categorized into \textit{upset, sadness, disappointment and regret}, \textit{shame} has been categorized into \textit{shame and embarrassed}. This breakdown into more fine-grained emotion labels helps us to move away from a fixed set of emotion labels and capture more nuanced emotions in the input dataset. 

For the generated emotional reasoning, we manually evaluate 100 samples from each dataset to assess the quality of the top-3 emotion labels and explanations. Examples of the generated explanations are listed in Table \ref{tab:isear_examples}.
Since there exists no ground truth for these tasks, we evaluate each output using the set of questions below:
\begin{enumerate}
    \item Does this label correctly represent the emotion expressed by the input text?
    \item Is this label more appropriate than the gold emotion label for the input text?
    \item Is the emotional reasoning correct?
    \item Is the reasoning grammatically correct?
    \item Is the reasoning complete?
    
\end{enumerate}

Each question can be answered by either 1 (Yes), 2 (Maybe) or 3 (No). For question 2, the answer Maybe corresponds to "New emotion label is same as gold label". Thus, it helps us to identify the number of times the generated labels are better than the gold label. Questions 4 and 5 are included to evaluate the quality of the generated texts.

\begin{figure}[!h]
\centering
    \includegraphics[width=0.95\linewidth]{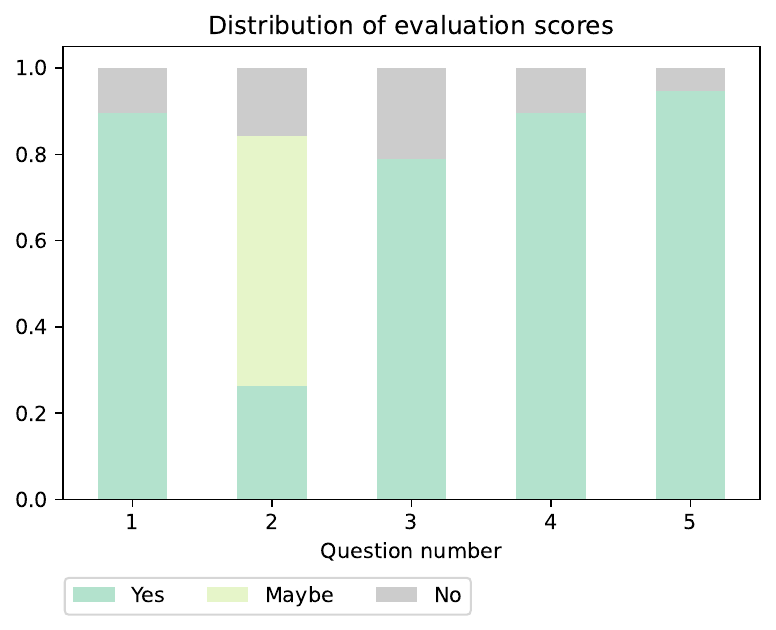}
    \caption{Human evaluation of top-3 generated emotional reasoning in ISEAR dataset. Plot shows the distribution of scores for questions 1-5.}
    \label{fig:isear_evals}
\end{figure}

Fig. \ref{fig:isear_evals} summarizes the scores for the evaluation of the generated outputs for the ISEAR dataset. Over 89\% of the generated labels correctly expresses the emotion of the input text. For question 2, 57\% of the generated labels are noted to be similar to the gold label and 26\% of them are rated to be more appropriate than the gold label.
The emotional reasonings are correct for over 80\% of the times. It is to be noted that only the top-3 explanations are manually evaluated here. In multiple of the remaining 20\% cases we find a more plausible reasoning within the unexamined explanations.

\begin{figure*}[t]
\centering
    \includegraphics[width=0.9\linewidth]{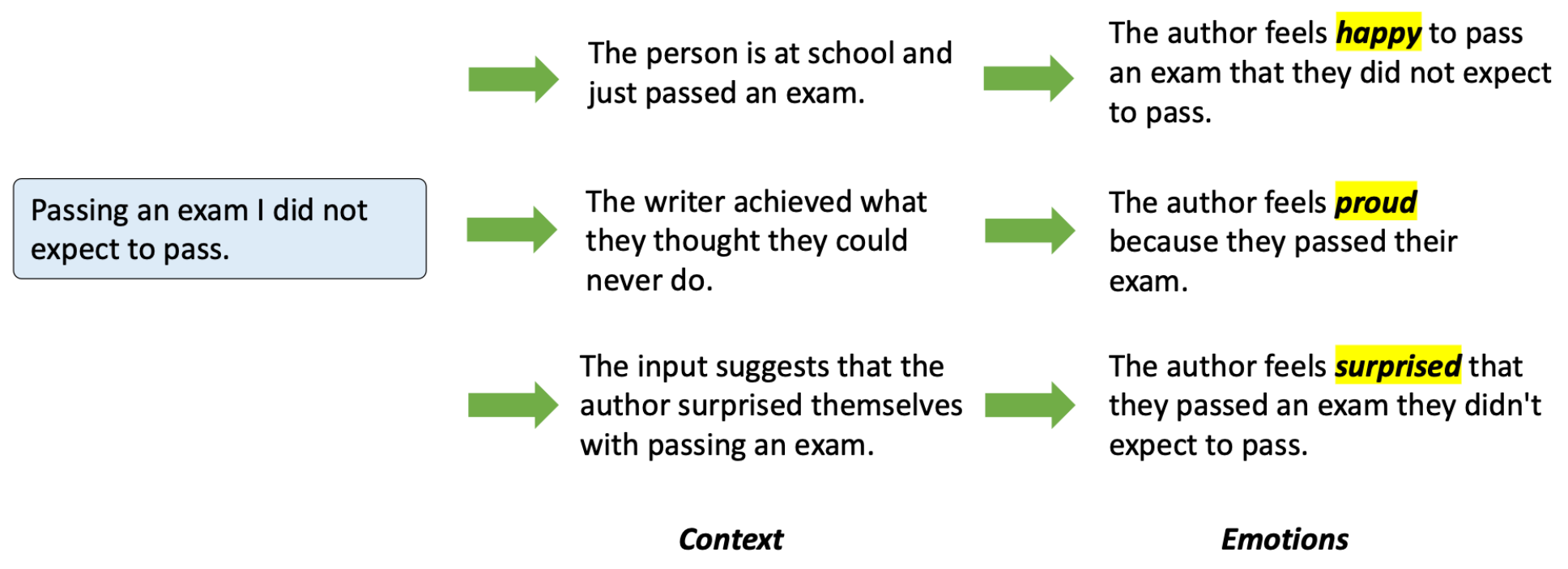}
    \caption{Example from ISEAR illustrates how different contexts result in a distinct CoT emotion reasoning.}
    \label{fig:context}
    \vspace{-0.2in}
\end{figure*}

\subsection{Significance of Context Generation}
We analyze the complete set of generated outputs for the ISEAR dataset before extracting the top-3 emotion labels from them. We carry out this experiment to demonstrate the significance of the context generation step in our approach. We observed several instances where different contexts result in distinct CoT reasonings and, consequently, result in different emotion labels (Fig. \ref{fig:context}). This observation aligns with our initial claim that context plays an important part in generating the appropriate emotion label for a particular dataset or domain.

We repeat the experiment of emotion reasoning generation without the incorporation of context. We observe a drastic decrease in the quality of the CoT reasoning process. The model mostly replicates text to produce a reasoning from the input in the absence of an intermediate context.

\subsection{Error Analysis}
We analyze the 20 examples that were rated to have incorrect emotional reasoning. Errors in reasoning occur due to three major reasons: (1) the LLM is unable to capture a multiple clauses in the input text, (2) the LLM outputs the emotion label only with no explanation, (3) the input text contains some other emotion word which confuses the LLM. Examples of errors have been listed in Table \ref{tab:errors}.

\begin{table*}[!t]
\small
    \centering
    \begin{tabular}{p{0.35\linewidth}|p{0.35\linewidth}|p{0.2\linewidth}}
    \hline
    Text & Reasoning & Error Category \\
    \hline
    When I realized that I was directing the feelings of discontent with myself at my partner and this way was trying to put the blame on him instead of sorting out my own feelings. & The author feels blame because they feel discontent with themselves and want to put the blame on their partner. The final emotion label is \textbf{blame}. & Emotion label in input \\
    \hline
    After my girlfriend had taken her exam we went to her parent's place.
    & The author feels happy because their girlfriend took her exam. The final emotion label is \textbf{happy}. & Multiple reasons \\
    \hline
   When one has arranged to meet someone and that person arrives late, in the meantime one starts thinking about all that could have gone wrong e.g a traffic accident.
    & The author feels \textbf{panicky}. & No explanation \\
    \hline
    \end{tabular}
    \caption{Examples of errors in generated explanations.}
    \label{tab:errors}
\end{table*}

\begin{table*}[!t]
\small
    \centering
    \begin{tabular}{p{0.3\linewidth}|p{0.06\linewidth}|p{0.07\linewidth}|p{0.13\linewidth}|p{0.33\linewidth}}
    \hline
    Input & Gold Label & Predicted Label & Other Generated Labels & Reasoning \\
    \hline
    When I think about the short time that we live and relate it to the periods of my life when I think that I did not use this short time. & sadness & sadness & regret & 
    The author feels sad when they think of how they wasted their time in the past. The final emotion label is \textbf{sad}. \newline
    The author feels regret because the author realizes that they did not use the short time that they were alive in the most efficient way possible. The final emotion label is \textbf{regret}.
     \\ \\
     When I think that I do not study enough.  After the weekend I think that I should have been able to have accomplished something during that time. & guilt & guilt & nervousness, \newline disappointment & 
    The author feels guilt because they failed to do what they should have done in the weekend. The final emotion label is \textbf{guilt}. \newline
    The author feels nervousness because they need to study a lot to maintain a good grade. The final emotion label is \textbf{nervousness}. \newline
    The author feels disappointment because the author was not able to accomplish something that was important to him. The final emotion label is \textbf{disappointment}. 
    \\ \\
     When I found out that I had passed the last two exams by a margin of three marks. & joy & joy & happiness, \newline relieved & 
    The author feels happy because they passed two exams. The final emotion label is \textbf{happiness}. \newline
    The author feels relieved because he passed exams by a margin of three marks. The final emotion label is \textbf{relieved}. 
    \\ \\
    
    I feel awful, and it's way too freaking early. Now off to leadership highschool...  & joy & sadness & sad, \newline awful & 
    The author feels sad because they don't like going to school and they don't like the subject they have to study for. The final emotion label is \textbf{sad}. \newline
    The author feels awful because they feel like going to leadership school is not fun. The final emotion label is \textbf{awful}.
    \\ \\
    I have a package at the post office. Can't think what could be in it.  I don't remember internet shopping while drinking. & surprise & surprise & confused & 
    The author feels surprise because he does not remember the purchase. The final emotion label is \textbf{surprised}. \newline
    The author feels confusion because the author is unsure what was inside the package. The final emotion label is \textbf{confused}.
    \\

    \hline
    \end{tabular}
    \caption{Examples from the updated datasets illustrating the other relevant emotion labels generated and the explanations generated for each of them.}
    \label{tab:isear_examples}
\end{table*}

\section{Related Work}

\subsection{Emotion Detection from Text}
Current state-of-the-art emotion detection systems can be largely categorized into two categories: domain specific fine-tuned models \citep{ma2019pkuse, chiorrini2021emotion} and zero-shot systems that rely on sentence embeddings and textual entailment models \citep{zhangkumjornZeroShot, olah2021cross, zero_shot_social_2022}.
In spite of impressive performance across benchmarks these systems have limitations. They are extremely rigid when it comes to adapting to a new set of emotion labels or domains. Moreover, they are black box systems that only produce a set of emotion scores for an input text.
This leads us to the tasks of emotion cause extraction and emotion cause-pair extraction that focus on identifying the reasons behind the detected emotions \citep{lee-etal-2010-text, chen2018hierarchical, xia-ding-2019-emotion}. Although these tasks can extract word or clause level emotion triggers, they are constrained to identifying causes solely within the input text. Recent work by \citealp{yang2023interpretable} defines the task of \textit{Emotional Reasoning} for mental health analysis as a combination of emotion recognition from conversations and causal emotion entailment. We argue that these classification tasks do not capture the intuitive step-by-step reasoning process that humans employ for emotional reasoning.

\subsection{Reasoning using LLMs}
Recently, pre-trained LLMs like GPT-3 \citep{brown2020language}, OPT \citep{zhang2022opt}, T5 \citep{raffel2020exploring}, ChatGPT \citep{ouyang2022training} have become extremely popular for various natural language processing tasks. With the introduction of Chain-of-Thought prompting techniques \citep{wei2022chain, narang-self}, the reasoning capabilities of these LLMs have been explored for commonsense, arithmetic and science question answering tasks \citep{rajani-etal-2019-explain, lin-etal-2020-birds,  lu2022learn, liu-etal-2022-generated, fei-etal-2023-reasoning}. We rely on these works to explore the task of emotional reasoning using LLMs.

\subsection{Context Generation for QA Tasks}
Several recent works have also shown the importance of incorporating LLM generated text into prompts for QA tasks. Augmenting the input question with knowledge generated by the LLM has improved the performance of commonsense reasoning tasks \citep{liu-etal-2022-generated}. Generation of context to aid in closed-book QA has outperformed existing fine-tuned models \citep{su-etal-2023-context}. We follow these works to show that the incorporation of context is essential for the task of emotional reasoning and adapt it to multiple domains.



\section{Conclusion}
In this paper, we present a two-step generative approach to the task of emotion analysis. We introduce the task of step-by-step emotional reasoning to explain the chain-of-thought leading to the emotion label. We use existing emotion detection datasets to evaluate our approach and will release the datasets updated with additional emotion labels and their explanations. 
Future work will include the use of LLMs in multi-step emotional CoT reasoning for longer texts and dialogues. This would be a step towards building more human-like empathetic conversational assistants with potential applications across various domains.

\section{Acknowledgements}
Details of research grants funding this work have been withheld to maintain anonymity. It will be included in the camera ready version.

\section{Limitations and Ethical Considerations}
In this work, we introduce a novel generative approach to emotion detection from text and compare against popular zero-shot methods and prompting techniques. We do not compare against existing fine-tuned emotion detection models owing to the fact that they work only over fixed sets of emotion labels. Our work focuses more on a generalized approach to the task that can be adapted to specific domains using the context generation step.
It is crucial to manually write the few-shot prompt for every dataset after careful consideration of the dataset creation process. It is to be noted that throughout our work we experiment with multiple prompts and report the most effective ones in the paper.
Further, the use of smaller LLMs are not able to generate any relevant context or perform consistent reasoning. This also makes our system susceptible to common biases or hallucinations that are prevalent in these LLMs \citep{rawte2023survey}.

\nocite{*}
\section{Bibliographical References}\label{sec:reference}

\bibliographystyle{lrec-coling2024-natbib}
\bibliography{lrec-coling2024-example}



\end{document}